\documentclass[conference]{IEEEtran}
\usepackage{cite}
\usepackage{amsmath,amssymb,amsfonts}
\usepackage{algorithmic}
\usepackage{algorithm} 
\usepackage{graphicx}
\usepackage{textcomp}
\usepackage{xcolor}
\usepackage{subcaption}
\newtheorem{definition}{Definition} 
\newtheorem{proof}{Proof} 
\usepackage{wrapfig}
\usepackage{enumitem}
\usepackage{url}

\begin{document}

\title{Fair Algorithms for Hierarchical Agglomerative Clustering\\
}

\author{\IEEEauthorblockN{Anshuman Chhabra and Prasant Mohapatra}
\IEEEauthorblockA{\textit{Department of Computer Science} \\
\textit{University of California, Davis}\\
Davis, California, USA \\
\texttt{\{chhabra, pmohapatra\}@ucdavis.edu}}
}

\maketitle

\begin{abstract}
  Hierarchical Agglomerative Clustering (HAC) algorithms are extensively utilized in modern data science, and seek to partition the dataset into clusters while generating a hierarchical relationship between the data samples. HAC algorithms are employed in many applications, such as biology, natural language processing, and recommender systems. Thus, it is imperative to ensure that these algorithms are \textit{fair}-- even if the dataset contains biases against certain \textit{protected groups}, the cluster outputs generated should not discriminate against samples from any of these groups. However, recent work in clustering fairness has mostly focused on center-based clustering algorithms, such as \textit{k-median} and \textit{k-means} clustering. In this paper, we propose fair algorithms for performing HAC that enforce fairness constraints 1) irrespective of the distance linkage criteria used, 2) generalize to any natural measures of clustering fairness for HAC, 3) work for multiple protected groups, and 4) have competitive running times to vanilla HAC. Through extensive experiments on multiple real-world UCI datasets, we show that our proposed algorithm finds \textit{fairer} clusterings compared to vanilla HAC as well as the only other state-of-the-art fair HAC approach. 
\end{abstract}


\section{Introduction}\label{sec:intro}
Hierarchical Agglomerative Clustering (HAC) refers to a class of greedy unsupervised learning algorithms that seek to build a hierarchy between data points while clustering them in a bottom-up fashion. HAC algorithms are widely utilized in modern data science-- such as in genetics \cite{pagnuco2017analysis}, genomics \cite{pollard2005cluster}, and recommendation systems \cite{merialdo1999clustering}. These algorithms also possess two distinct advantages over non-hierarchical or \textit{flat} clustering algorithms: 1) they do not require the number of clusters to be specified initially, and 2) they output a hierarchy among all samples in the dataset.

The machine learning community has realized the importance of designing \textit{fair} algorithms. Traditional machine learning algorithms do not account for any biases that may be present (against certain minority groups) in the data, and hence, may end up augmenting them. Machine learning is being increasingly utilized in societal applications such as loan defaulter prediction \cite{ZHU2019503}, recidivisim rate prediction \cite{duwe2017out}, and many more. Owing to the sensitive nature of these applications and their far reaching impact on human lives, ensuring that these algorithms are fair becomes of paramount importance. However, work in designing fair clustering algorithms has mostly been focused either on the center-based or graph-based clustering objectives like \cite{bercea2018cost,chierichetti2017fair,backurs2019scalable,chen2019proportionally,kleindessner2019guarantees,anagnostopoulos2019principal}. Despite all the advantages of HAC algorithms, there has been little work that proposes fair variants to HAC. In this paper, we seek to bridge this gap by making the following contributions:
\begin{itemize}[wide]
    \item The proposed fair HAC algorithm (Section \ref{section3}) works for multiple protected groups, and we provide results on all the widely used linkage criteria (single-linkage, average-linkage, complete-linkage) for real datasets (Section \ref{section4}). Our algorithm can also generalize to any natural analytical notions of fairness for HAC (Section \ref{section3}).
    \item Our algorithm achieves an asymptotic time complexity of $\mathcal{O}(fn^3)$ ($f$ is the number of protected groups), which is comparable to $\mathcal{O}(n^3)$ for traditional HAC as $f$ is a small number for most applications (Section \ref{section3}).
    \item We provide experimental analysis for different \textit{fairness costs} for our algorithm, and show that it is more \textit{fair} than vanilla HAC (Section \ref{section4_2}), and the only other fair HAC algorithm proposed by \cite{ahmadian2020fair} (Section \ref{section4_3}). Also, the algorithm of \cite{ahmadian2020fair} cannot generalize to any arbitrary notion of fairness.
    \item We undertake an experimental analysis of cluster quality between our proposed algorithm, vanilla HAC, and the algorithm of \cite{ahmadian2020fair}, using the Silhouette Coefficient \cite{ROUSSEEUW198753} and find that our algorithm generally obtains higher quality clusters with improved fairness (Section \ref{section4_4}). 
\end{itemize}

\noindent\textbf{Motivating Example.} We provide an example similar to the job shortlisting example given in \cite{deepak2020whither}, but tailor it to shortlisting households/individuals for bank credit promotions. A dataset, e.g., the \texttt{creditcard} dataset \cite{yeh2009comparisons}, is used by the marketing division of a bank to reach out to prospective customers and offer them loans and available credit opportunities. The dataset contains information on the potential customer's age, education level, weekly work hours, and capital gains per month. The bank utilizes a hierarchical clustering algorithm to find target audiences for promotional offers, using the aforementioned attributes as input features. On running the algorithm, they obtain hierarchical \textit{clusters} of people. The bank then chooses an appropriate $k$ number of clusters based on available credit offers. A small number of clusters are shortlisted for a particular promotion using chosen metrics, (e.g. education and wages-earned) to represent clusters and select a few of them. However, people-of-color (POC) and women tend to earn lower wages than Caucasian males\footnote{\url{https://www.pewresearch.org/fact-tank/2016/07/01/racial-gender-wage-gaps-persist-in-u-s-despite-some-progress/}}, and that POC face more adversities that lead to disparities in their education level as opposed to white demographics\footnote{\url{https://www.brookings.edu/blog/brown-center-chalkboard/2016/06/06/7-findings-that-illustrate-racial-disparities-in-education/}}. Considering these facts on the racial education divide and the wage gap, a clustering algorithm using these attributes will inherently group white households and men as better candidates. As a result, this marketing clustering algorithm has \textit{disparate impact} on POC as well as women. In this case, race/ethnicity and gender can be considered as \textit{protected groups} and the objective of the fair hierarchical clustering algorithm is to ensure that each group gets a certain minimum representation. Moreover, no group should be overwhelmingly preferred, meaning there should be a cap on the maximum representation allowed. This example demonstrates how \textit{bounded representation} mitigates disparate impact in the context of clustering. Our goal is to develop a hierarchical clustering algorithm to output trees that abide by this notion of fairness.

The rest of the paper is organized as follows: Section \ref{section2} discusses related work in the field, Section \ref{section3} details our proposed algorithm for performing fair HAC, Section \ref{section4} describes our results on real data, and Section \ref{section5} concludes the paper and discusses the scope of future work.

\section{Related Works}\label{section2}
Recently, many works have focused on providing clustering algorithms with fairness guarantees \cite{chhabra2021overview}. However, most of this work has studied center-based clustering (such as k-means, k-center, and k-median) \cite{bercea2018cost,chierichetti2017fair,backurs2019scalable,bera2019fair, chhabra2022fair}, and spectral methods \cite{kleindessner2019guarantees, chhabra2022fair}. This line of work seeks to impose some \textit{fairness} constraints along with the original clustering distance based objective. Chierichetti et. al were the first to propose fair clustering via \textit{fairlet decomposition} for k-center and k-median clustering \cite{chierichetti2017fair} using a fairness definition based on disparate impact \cite{feldman2015certifying}. They ensured that in the case of two protected groups, each cluster had points of both groups, measured using a metric known as \textit{balance}. Numerous works have followed that improve upon these ideas such as better approximation rates \cite{backurs2019scalable}, \cite{bercea2018cost}, allowing for multiple protected groups \cite{bercea2018cost}, extending to other clustering objectives \cite{kleindessner2019guarantees}.

Other than our work and Ahmadian et al's Fair Hierarchical Agglomerative Clustering (AFHAC)\cite{ahmadian2020fair}, no work has investigated fairness in the context of hierarchical clustering so far. The key difference between our work and AFHAC is that we work with the greedy HAC algorithms and AFHAC works with hierarchical clustering objectives. In particular, Ahmadian et al's work considers objectives for hierarchical clustering that have been proposed following Dasgupta's seminal work \cite{dasgupta2016cost}, such as \cite{moseley2017approximation}, \cite{cohen2019hierarchical}. However, none of these objectives are approximated well-enough by any general distance linkage criteria that are typically used in HAC (except for average-linkage). Moreover, greedy HAC algorithms, despite being ad-hoc and heuristic approaches, are widely utilized in many application areas like biological sciences. For these reasons, we wanted to ensure fairness for these algorithms specifically, and provide a fair variant to the HAC problem, irrespective of the choice of distance linkage criteria. In the results section, we compare the fairness achieved by our algorithms to the AFHAC algorithms (for both the \textit{revenue} \cite{moseley2017approximation} and \textit{value} \cite{cohen2019hierarchical} hierarchical clustering objectives) on a number of datasets. Through extensive experiments we find that our proposed algorithm achieves better fairness than AFHAC. 

\section{Performing Fair HAC}\label{section3}
\looseness-1 First, we need to define the vanilla HAC process formally. Let $X \in \mathbb{R}^{n \times m}$ be our dataset. Then the HAC on $X$ denoted by $HC(X)$ is a hierarchical partitioning of $X$ that is represented by a binary tree $T$ (also called a \textit{dendogram}), where each level of $T$ represents a set of disjoint merges between subclusters. Each node of $T$ at any level represents a subcluster of points. An HAC algorithm first considers each of the $n$ samples of $X$ to be singleton subclusters, and then chooses sets of two subclusters to merge together. A point from $X$ can only belong to any one subcluster at any particular level. The lowest level of $T$ are leaves, and comprise of all the $n$ points of $X$. The root of $T$ is a single node/cluster that contains all of $X$. Let $C_1, C_2, ..., C_s$ be the subclusters at any level of $T$. Then $C_1 \cup C_2 \cup ... \cup C_s = X$. The choice of which two subclusters should be merged is made by finding two subclusters $C_i$ and $C_j$ such that they minimize a \textit{linkage criterion} denoted by $D(C_i,C_j)$. There are many linkage criteria that can be used. For example, single-linkage is defined as $D(C_i,C_j) = \min_{x_i \in C_i, x_j \in C_j} d(x_i,x_j)$ and complete-linkage is defined as $D(C_i,C_j) = \max_{x_i \in C_i, x_j \in C_j} d(x_i,x_j)$. In this paper we assume $d(x,y)$ is the Euclidean distance between two points $x$ and $y$, but other distance metrics can also be used.

Next, we need to define our proposed notion for fairness. In this work we consider data points to only belong to one protected group, that is, \textit{multiple} assignments to many protected groups for the same point are not allowed. In most works of fairness in clustering, the notion of \textit{balance} \cite{chierichetti2017fair} is used to ensure that clusters contain each protected group in desired proportions. We work with a similar notion, but frame this metric as a \textit{fairness cost} that will naturally align with fair hierarchical clustering. We also introduce an input parameter called the \textit{ideal proportion} ($\phi_g$) for each group $g$. This value is the desired proportion of points for each group in each cluster. Thus the ideal proportion of points from protected groups in each cluster can be set depending on the application context. In accordance with the \textit{disparate impact} doctrine, we let the \textit{ideal proportion} of each protected group be the proportion of the group in the entire dataset $X$ \cite{bera2019fair}. Thus, for a protected group with $s$ members, the ideal proportion would be $s/n$. This can be easily altered for when all groups should be equally balanced in each cluster-- if we have $f$ protected groups, we can replace the \textit{ideal proportion} for each group with $1/f$. Finally, the only other existing work on fair hierarchical clustering \cite{ahmadian2020fair}, employs a definition of fairness where the fairness of all protected groups is upper-bounded by the same value. Our definition is thus more general, since we can define different ideal proportions for each group.

\begin{definition}{\textbf{($\alpha$-Proportional Fairness)}}
Let $F \in \mathbb{R}^{f \times n}$ be the set of all protected groups where each protected group $g\in F$ is $\{0,1\}^{n}$. Thus, if data sample $j$ from $X = \{x_i\}_{i=1}^{n}$ belongs to a particular group $g$ then at the $j$-th index $g$ contains a $1$, otherwise a $0$. Moreover, a cluster $C = \{x_i | i \in I\}$ where $C \subset X$, and $I$ is the index set containing indices of the points in $X$ which belong to cluster $C$. The proportion of group $g$ members in $C$ is denoted by $\delta_g^C = \frac{1}{|C|}\sum_{x_i \in C} g_i$ and the ideal proportion $\phi_g = \frac{1}{n}\sum_{x_i \in X} g_i$. Then $\alpha$-Proportional Fairness for cluster $C$ and protected group $g$ is maintained if the following condition holds: $|\delta_g^C - \phi_g| \leq \alpha$.
\end{definition}

\begin{definition}{(\textbf{Max Fairness Cost (MFC)})} Let $HC(X)$ be the output of some hierarchical clustering on $X$. Then the fairness cost on some level with $k$ clusters of the $HC(X)$ tree measures how close each cluster of points at this level (denoted by $C_i$, where $i = \{1,2,..,k\})$ is to the ideal proportion $\phi_g$ for each protected group $g$ in $F$. Mathematically, the Max Fairness Cost can then be defined as: $\max_{i \in [k], g \in F}|\delta_g^{C_i} - \phi_g|$.
\end{definition}

Minimizing the MFC minimizes the cumulative maximum deviation of groups from the ideal proportion, for all $k$ clusters and all $g$ groups. As a result, all clusters have proportions of all protected groups as close as possible to the ideal proportions. 

We also include analysis for the balance metric. Below, we define the multiple-group version of balance, initially defined by \cite{chierichetti2017fair} for the 2-group case, and then generalized by \cite{bera2019fair}:

\begin{definition}{(\textbf{Multi-Group Balance \cite{bera2019fair}})} Following from the notation established above, the balance of a clustering can be defined as: $\min_{i\in [k]}\min\{\frac{\delta_g^{C_i}}{\phi_g}, \frac{\phi_g}{\delta_g^{C_i}}\}$, $\forall g\in F$.

\end{definition}

Balance needs to be maximized to improve fairness. The obtained balance will always lie between $[0,1]$, where a value of $0$ is a completely unbalanced clustering (least fair) and  a value of $1$ is a perfectly balanced clustering (most fair). In the next subsection, we delineate how fairness metrics such as the MFC or balance can be optimized for HAC using our proposed algorithm.

\subsection{Fair HAC: The FHAC Algorithm}
Thus, the goal for our fair algorithm is to minimize the fairness cost by ensuring that at each level of the tree, we have at least maintained some $\alpha'$-Proportional Fairness where $\alpha'$ is some constant. We run the algorithm till some $k$ clusters are remaining, and return after that. To enforce these constraints, we keep tightening the bound on proportional fairness while simultaneously loosening the bound on minimizing distance, as we start getting closer to $k$ clusters. The greedy Fair Hierarchical Agglomerative Clustering (FHAC) algorithm is described as Algorithm 1 (which works irrespective of the choice of linkage criteria). Algorithm 1 resembles vanilla HAC with some key distinctions that allow it to be fairer. The key difference is foregoing the minimum distance linkage criterion constraint to allow for the selection of fairer clusters that can be merged, leading to a fairer output tree. It is important to note that throughout, we maintain a distance matrix $\mathcal{D} \in \mathbb{R}^{n \times n}$ (line 2) between clusters to improve the runtime of the algorithm; this can be done by computing the distances using the linkage criterion provided as input. Also, $d_{\min}$ (line 5) signifies the initialization of the distance (computed using the linkage criterion) between the clusters chosen to be merged for this level. 

Algorithm 1 functions as follows: it tightens the fairness constraint (and loosens the distance constraint) as we keep constructing the clustering tree from bottom to top. The fairness constraint tightening is achieved using the function $\mathcal{Z}_{\alpha}: \mathbb{R} \rightarrow \mathbb{R}$ and the distance constraint loosening is achieved using the $\mathcal{Z}_{\beta}: \mathbb{R} \rightarrow \mathbb{R}$ function. $\mathcal{Z}_{\alpha}$ and $\mathcal{Z}_{\beta}$ are both monotonically increasing functions and are parameterized appropriately for the dataset $X$. We compute the actual constraint bounds $\alpha$ and $\beta$ using $\mathcal{Z}_{\alpha}(|C| - k)$ (line 6) and $\mathcal{Z}_{\beta}(n - |C|)$ (line 7), respectively, where $C$ in each iteration of the while loop (line 3) denotes the current state of clusters at some level of the tree. Intuitively-- if both functions are monotonically increasing, then $\alpha$ reduces as we get closer to $k$ clusters, tightening the fairness constraint, whereas $\beta$ increases as we get closer to $k$ clusters, loosening the distance constraint. 

\begin{algorithm}[H]
\fontsize{9}{9}\selectfont
\caption{Proposed FHAC Algorithm}
 \textbf{Input:} $X$, $F$, $k$, $D(.,.), \mathcal{Z}_{\alpha}: \mathbb{R} \rightarrow \mathbb{R}, \mathcal{Z}_{\beta}: \mathbb{R} \rightarrow \mathbb{R}$\\ 
 \textbf{Output:} Fair HAC tree $T_{fair}$ 
 \begin{algorithmic}[1]
 \STATE \textbf{set} $C \leftarrow X$
 \STATE \textbf{compute} $\mathcal{D}_{n_1,n_2} = D(n_1,n_2), \forall (n_1,n_2) \in X \times X$ 
 \WHILE{$|C| \geq k$}
 \STATE  $P_g = 0$, $\forall g \in F$
 \STATE  $d_{\min} \leftarrow \infty$
 \STATE  $\alpha \leftarrow \mathcal{Z}_{\alpha}(|C| - k)$
 \STATE  $\beta \leftarrow \mathcal{Z}_{\beta}(n - |C|)$
 
 \FOR{each $(c_i,c_j) \in C \times C$,  \textit{s.t.} $c_i \neq c_j$} 
 
 \FOR{each $g \in F$}
 
 \STATE $\delta_g^{c_i + c_j} \leftarrow \frac{\delta_g^{c_i}|c_i| + \delta_g^{c_j}|c_j|}{|c_i| + |c_j|}$
 
 \STATE \textbf{if} $|\delta_g^{c_i + c_j} - \phi_g| \leq \alpha$ \textbf{then} $P_g' = 1$, \textbf{else} $P_g' = 0$
 
 \ENDFOR
 
 \IF{$\sum_{g \in F}P_g' \geq \sum_{g \in F}P_g $}
 \IF{$d_{\min} + \beta > \mathcal{D}_{c_i, c_j}$}
 \STATE $d_{\min} \leftarrow \mathcal{D}_{c_i, c_j}$
 \STATE $P_g \leftarrow P_g', \forall g \in F$ 
 \STATE $(c_1^m, c_2^m) \leftarrow (c_i, c_j)$
 \ENDIF
 \ENDIF

 \ENDFOR
 
 \STATE \textbf{merge} $c_1^m \leftarrow c_1^m + c_2^m $
 \STATE \textbf{update} $C$ \textbf{with newly merged clusters}
 \STATE \textbf{recompute} $\mathcal{D}_{c_1,c_2} = D(c_1,c_2), \forall (c_1,c_2) \in C \times C$
 \STATE \textbf{update} $T_{fair}$ \textbf{with merge} 
 \ENDWHILE
 \STATE \textbf{return} $T_{fair}$
 
 \end{algorithmic}
\end{algorithm}

Unlike traditional HAC algorithms, we maintain the proportion of protected group $g$ members for each possible cluster pair to be merged (line 10). Another distinction is the $P_g \in \{0,1\}^f$ and $P_g' \in \{0,1\}^f$ vectors, that help us keep track of how many protected groups for a potential cluster merge have met the proportional fairness constraints. If $P_g' = 1$ we have met the proportionality condition for group $g$ in this iteration, otherwise $P_g' = 0$. $P_g$ is the same but maintains global state-- that is, it keeps track of the same condition but for the \textit{best} cluster merge pair found so far. In line 11, we check to see if we have met $\alpha$-Proportional Fairness if clusters $c_i$ and $c_j$ were to be merged, and then appropriately set the value for $P_g'$. Next, once we have done this for all the groups (lines 9-12), we compare the current cluster pair with the \textit{best} cluster merge pair found so far (line 13). If the current cluster pair is a better choice, we check whether the minimum distance constraint is improved (relaxed using $\beta$) on line 14, and update variables accordingly. Gradually we build up the HAC tree and return it as $T_{fair}$.

\textbf{Note.} As is evident, Algorithm 1 achieves an asymptotic time complexity of $\mathcal{O}(fn^3)$ which is comparable to vanilla HAC ($\mathcal{O}(n^3)$) since $f$ is usually small for real-world applications. We can further speed-up the clustering process by making locally optimal cluster merges if they meet the fairness (and distance) criteria, as opposed to searching the entire space for possible cluster pairs to merge. Furthermore, the benefit of employing Algorithm 1 for fair HAC, is that it can be utilized to minimize any cost function that is required for ensuring fairness of the hierarchical clustering process. In the previous section, we defined the MFC and balance for the level with $k$ clusters, which can be minimized by Algorithm 1 through appropriate parameterization of $\mathcal{Z}_{\alpha}$ and $\mathcal{Z}_{\beta}$. In the results section, we present experimental results for both these fairness metrics, demonstrating the generality of our approach.

\textbf{Calculation of hyperparameters.} For our empirical results we let $\mathcal{Z}_{\alpha}(x) = e^{\theta_1 x + \alpha_0}$ and $\mathcal{Z}_{\beta}(x) = e^{\theta_2 x + \beta_0}$ but any monotonic functions can be used. Thus, we have to estimate the parameters $\alpha_0$, $\beta_0$, $\theta_1$, and $\theta_2$ that minimize MFC or balance. For the results in the paper, we utilize a simple grid-search and then choose parameters based on the values obtained for the choice of fairness cost. Alternatively, any black-box hyperparameter search algorithm could be utilized for this purpose.

\subsection{Results on Toy Data}
We generate two-dimensional data from a uniform distribution ($[0,250]$) for our toy example and there are $25$ points in total ($n$) and $k = 4$. Clusters are denoted in Figure 1 using 4 colors (red, green, blue, and yellow).

\begin{figure}[htb!]
\centering
\begin{subfigure}{.4\textwidth}
   \centerline{\includegraphics[scale=0.33]{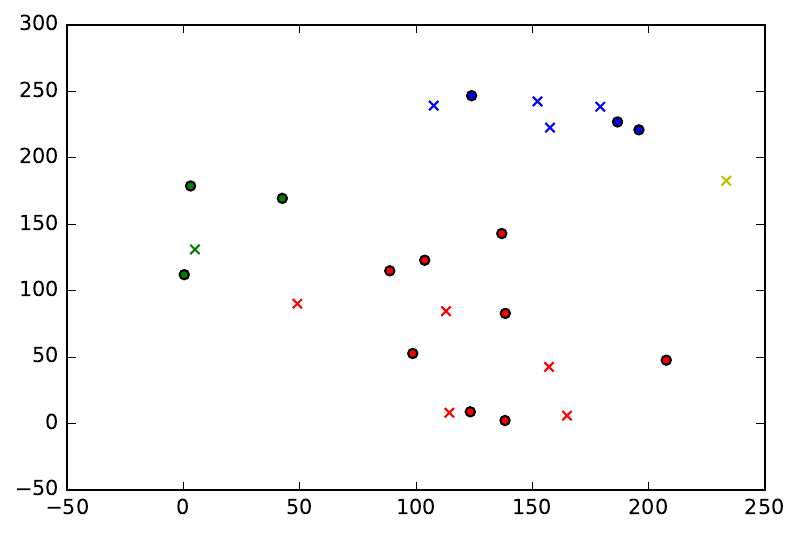}}
  \caption{Vanilla HAC clusters}
\end{subfigure}
\begin{subfigure}{.4\textwidth}
  \centerline{\includegraphics[scale=0.33]{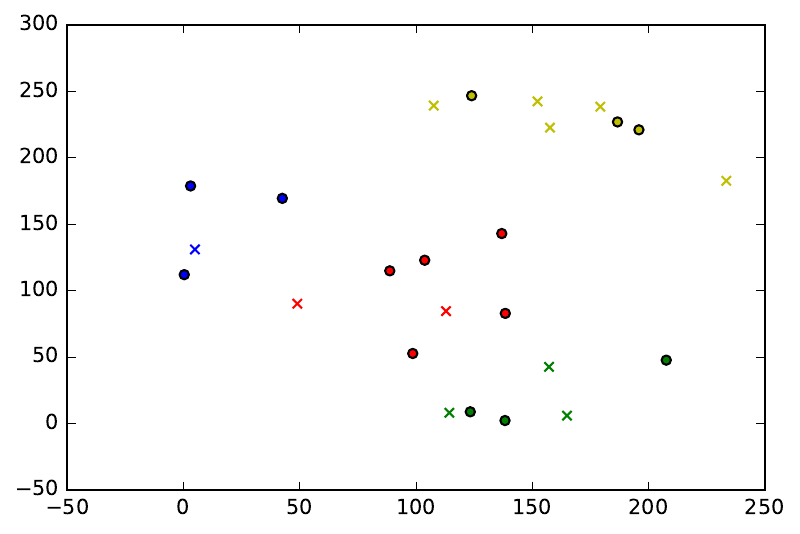}}
      \caption{FHAC clusters}
\end{subfigure}
\caption{Toy Dataset Results}
\end{figure}

There are two protected groups, denoted by $\circ$ and $\times$, with ideal proportions denoted by $\phi_{\circ} = 0.56$ and $\phi_{\times} = 0.44$, respectively. Therefore, a fair distribution of protected groups across clusters should be to (more, or less) balance them, as $\phi_{\circ} \approx \phi_{\times} \approx 0.5$. Also, $\mathcal{Z}_{\alpha}(x) = \theta_1 x + \alpha_0$ and $\mathcal{Z}_{\beta}(x) = \theta_2 x + \beta_0$. As mentioned above, we run FHAC (Algorithm 1) iteratively and estimate hyperparameters, and then compare how fair the final clusters are for vanilla single-linkage HAC and single-linkage FHAC. The values of the parameters are as follows: $\alpha_0 = 1.0, \theta_1 = 5.0, \beta_0 = 11.031, \theta_2 = 0.1826$. Moreover, MFC for vanilla HAC with single-linkage is $1.12$, and for FHAC with single-linkage is $0.38$. Thus, we find that the fairness achieved by our algorithm is much better, and we obtain proportionally fair clusters as a result. The clusters for traditional HAC are shown in Figure 1(a), and for our proposed FHAC are shown in Figure 1(b). Visually, it is easy to see that clusters in Figure 1(b) are more well-balanced than in Figure 1(a), where the red cluster has a large number of points, and the yellow cluster has only 1 point. Figure 1(b) instead has all points from each of the two protected groups distributed according to their ideal proportions.

\section{Results}\label{section4}
We utilize three real-world datasets to demonstrate the working of FHAC (Algorithm 1) and show that it computes \textit{fairer} (or equivalently fair) hierarchical clustering solutions compared to vanilla/traditional HAC (for average, complete, and single linkage criteria) in Section \ref{section4_2}, as well as the algorithms of Ahmadian et al for fair hierarchical clustering \cite{ahmadian2020fair} (for both value and revenue clustering objectives) in Section \ref{section4_3}. Furthermore we also analyze the quality of the fair clustering outputs obtained (using the Silhouette Coefficient \cite{ROUSSEEUW198753} clustering performance metric) between these algorithms in Section \ref{section4_4}, and find that our algorithms compute clusters of comparable quality. For Algorithm 1, we have $\mathcal{Z}_{\alpha}(x) = e^{\theta_1 x}$ and $\mathcal{Z}_{\beta}(x) = e^{\theta_2 x}$ for all experiments (that is we also set $\alpha_0 = \beta_0 = 0$).

\subsection{Datasets Used}\label{section4_1}

We utilize three datasets obtained from the UCI ML respository. We subsample datasets ($n=1000$) much like previous work \cite{chierichetti2017fair} \cite{abraham2019fairness}. We let merges occur up to $k=4$ (that is, we start $1000$ singleton clusters and merge them till we have $4$ clusters and we have obtained a hierarchical cluster tree) for all experiments and all datasets. These datasets are utilized in most research to evaluate fair clustering algorithms \cite{chierichetti2017fair}, \cite{bera2019fair}:
\begin{itemize}[wide]
    \item \textit{Creditcard Clients Dataset} \cite{yeh2009comparisons}: Consists of customers' default payments in Taiwan, denoted as \texttt{creditcard}. The features used are \textit{age, bill-amt 1 — 6, limit-bal, pay-amt 1 — 6}. The sensitive attribute used is \textit{education} with protected groups \textit{graduate school} ($\phi_0 = 0.25$), \textit{university} ($\phi_1 = 0.25$), \textit{high school} ($\phi_2 = 0.378$), \textit{others} ($\phi_3 = 0.122$). 
    \item \textit{Bank Marketing Dataset} \cite{moro2014data}: Consists of telephonic marketing campaigns of a Portuguese bank, denoted as \texttt{bank}. The features used are \textit{age, balance} and \textit{duration}. The sensitive attribute is \textit{marital status}, with protected groups \textit{married} ($\phi_0 = 0.334$), \textit{single} ($\phi_1 = 0.334$), \textit{divorced} ($\phi_2 = 0.332$).
    \item \textit{Census Income Dataset} \cite{kohavi1996scaling}: Contains 14 attributes of adults obtained via a 1996 census survey, denoted as \texttt{census}. The features used are \textit{age, education-num, final-weight, capital-gain, hours-per-week}. The sensitive attribute is \textit{sex} with protected groups \textit{male} ($\phi_0$$ = $$0.376$), \textit{female} ($\phi_1$$ = $$0.624$).
\end{itemize}


\subsection{Comparing FHAC (Algorithm 1) and Vanilla HAC}\label{section4_2}

\looseness-1 The fairness results (including parameter values) for running Algorithm 1 and vanilla HAC over all the aforementioned datasets for different linkage criteria for MFC as the fairness metric are shown in Table \ref{table1}. The lower the MFC, the more fair the obtained clustering is. The results when balance is selected as the fairness metric are shown in Table \ref{table2}. Here, the higher the balance, the \textit{fairer} the clustering obtained. Clearly, Algorithm 1 achieves \textit{fairer} solutions to vanilla HAC on real data.

\begin{table}[htb!]
\caption{(MFC Results) Vanilla HAC and FHAC (Ours)}\label{table1}
\begin{center}
\tabcolsep=0.10cm
\resizebox{0.42\textwidth}{!}{
\begin{tabular}{|c|c|c|c|c|c|c|c|}%
\hline
\textbf{Dataset} & \textbf{Linkage} & \textbf{$\theta_1$} & \textbf{$\theta_2$} & \textbf{MFC (Vanilla)} & \textbf{MFC (FHAC)}\\
\hline
\hline
\texttt{census} & Average & 0.001 & 0.001 & \textbf{0.0853} & \textbf{ 0.0853}\\
\hline
\texttt{census} & Complete & 0.001 & 0.001 & 0.1806 & \textbf{0.0853}\\
\hline
\texttt{census} & Single & 0.001 & 0.65 & 1.248 & \textbf{0.752}\\
\hline
\texttt{creditcard} & Average & 0.00005 & 0.005 & 1.2439 & \textbf{1.0}\\
\hline
\texttt{creditcard} & Complete & 0.00005 & 0.005 & 0.744 & \textbf{0.60}\\
\hline
\texttt{creditcard} & Single & 0.0075 & 0.5 & 1.5 & \textbf{ 0.778}\\
\hline
\texttt{bank} & Average & 0.0001 & 0.05 & 1.332 & \textbf{0.6679}\\
\hline
\texttt{bank} & Complete & 0.5 & 0.05 & 1.332 & \textbf{0.4457}\\
\hline
\texttt{bank} & Single & 0.001 & 0.65 & 1.332 & \textbf{0.6653}\\
\hline
\end{tabular}}
\end{center}
\end{table}
\vspace*{-5mm}

\begin{table}[htb!]
\caption{(Balance Results) Vanilla HAC and FHAC (Ours)}\label{table2}
\begin{center}
\tabcolsep=0.09cm
\resizebox{0.42\textwidth}{!}{
\begin{tabular}{|c|c|c|c|c|c|}
\hline
\textbf{Dataset} & \textbf{Linkage} & $\theta_1$ & $\theta_2$ & \textbf{Balance(Vanilla)} & \textbf{Balance(FHAC)}\\
\hline
\hline

\texttt{census} & Average & 0.001 & 0.001 &  \textbf{0.8865} & \textbf{0.8865}\\
\hline
\texttt{census} & Complete & 0.001 & 0.001 & 0.7599 & \textbf{0.8865}\\
\hline
\texttt{census} & Single & 0.0005 & 1.0 & 0.0 & \textbf{0.9385}\\
\hline

\texttt{creditcard} & Average & 0.05 & 0.1 & 0.0 & \textbf{0.427}\\
\hline
\texttt{creditcard} & Complete & 0.0005 & 1.0 & 0.0 & \textbf{0.7105}\\
\hline
\texttt{creditcard} & Single & 0.0005 & 1.0 & 0.0 & \textbf{0.7959}\\
\hline

\texttt{bank} & Average & 0.045 & 0.09 & 0.0 & \textbf{0.5567}\\
\hline
\texttt{bank} & Complete & 0.5 & 0.05 & 0.0 & \textbf{0.3327}\\
\hline
\texttt{bank} & Single & 0.0005 & 1.0 & 0.0 & \textbf{0.8099}\\
\hline

\end{tabular}}
\end{center}
\end{table}
\vspace*{-5mm}

\subsection{Comparing FHAC (Algorithm 1) and AFHAC}\label{section4_3}

\begin{figure*}[htb!]
\centering
\begin{subfigure}{.3\textwidth}
   \centerline{\includegraphics[scale=0.17]{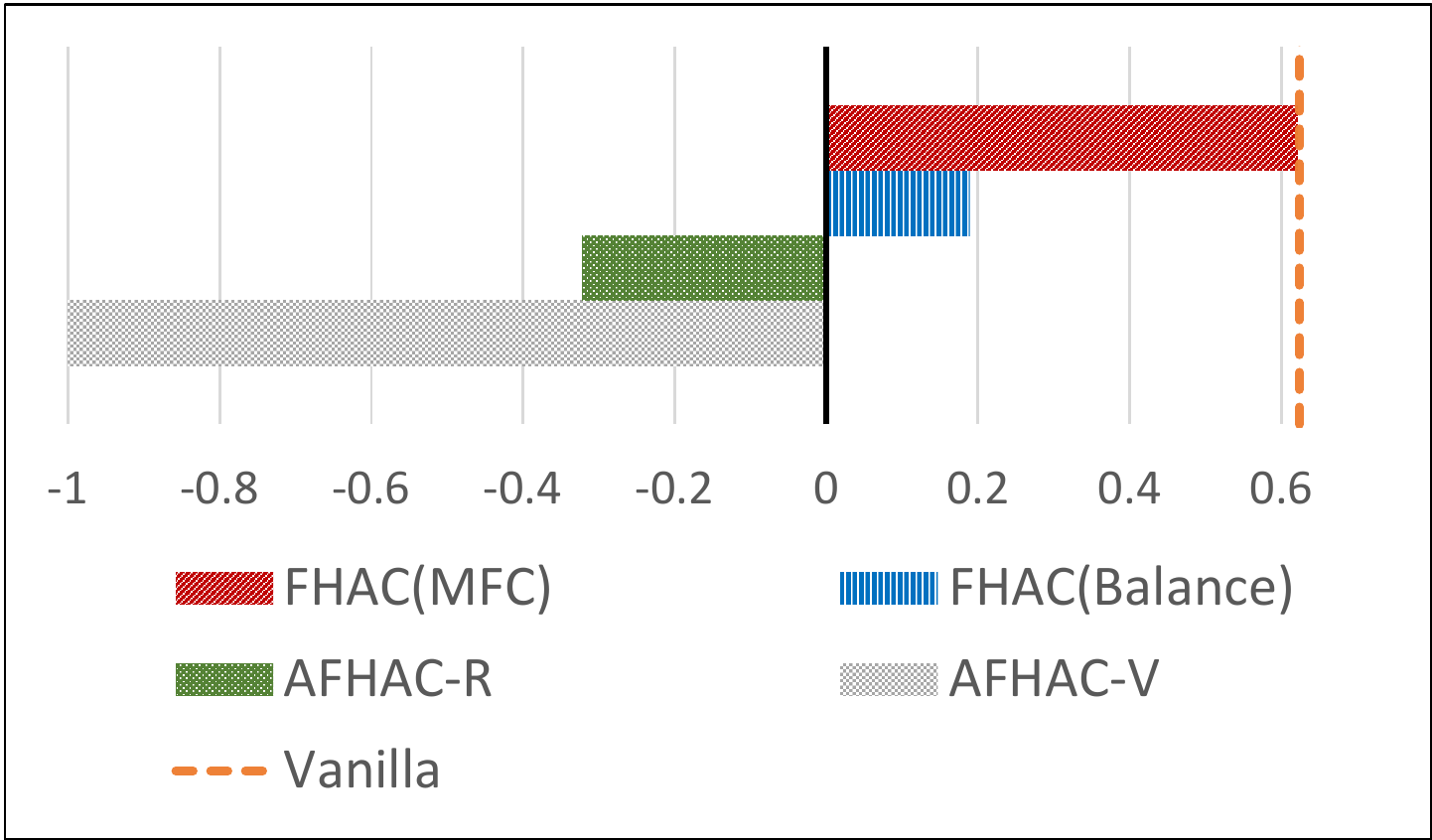}}
  \caption{Silhouette Scores for \texttt{creditcard}}
\end{subfigure}
\begin{subfigure}{.3\textwidth}
  \centerline{\includegraphics[scale=0.17]{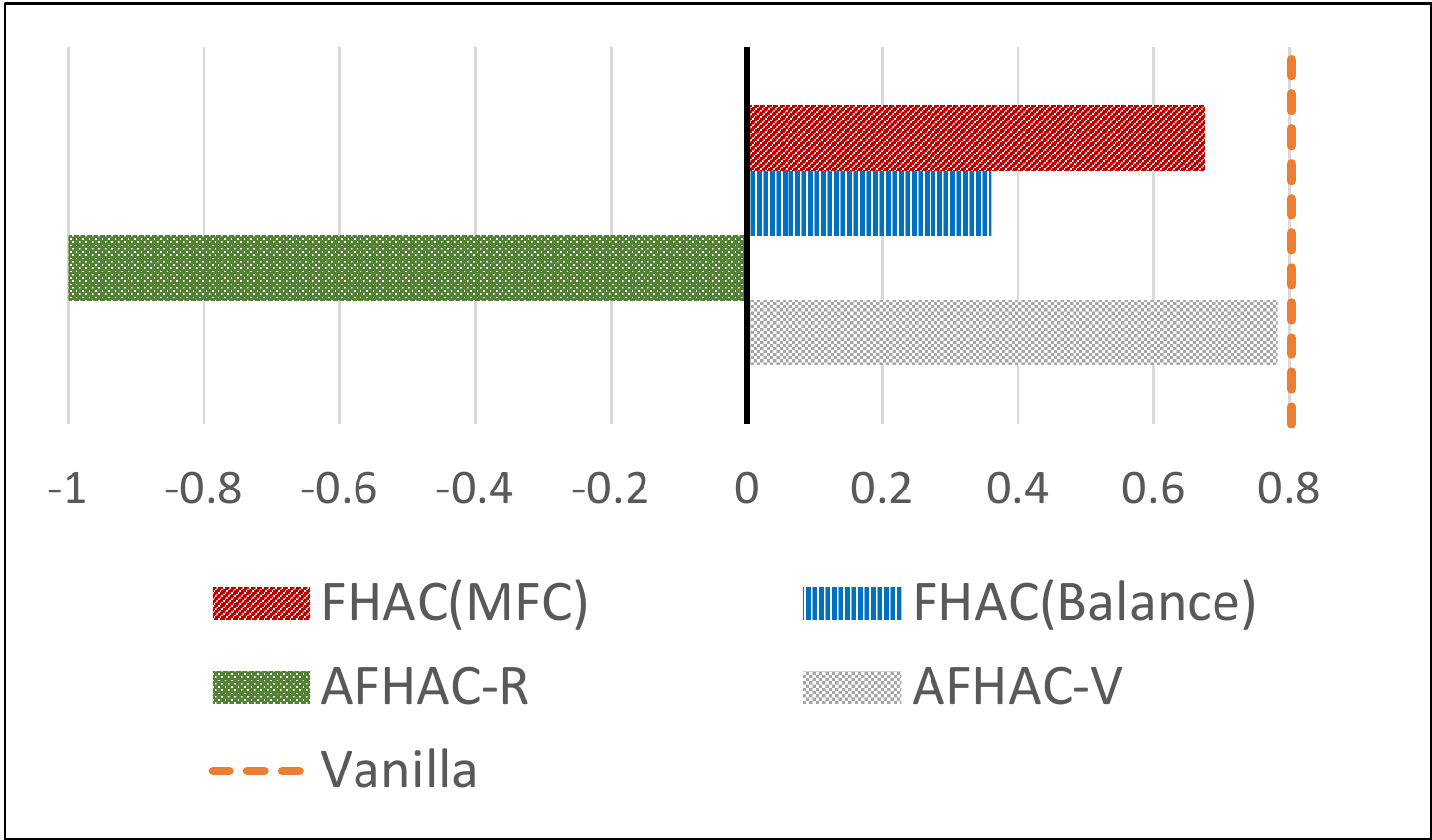}}
      \caption{Silhouette Scores for \texttt{bank}}
\end{subfigure}
\begin{subfigure}{.3\textwidth}
  \centerline{\includegraphics[scale=0.17]{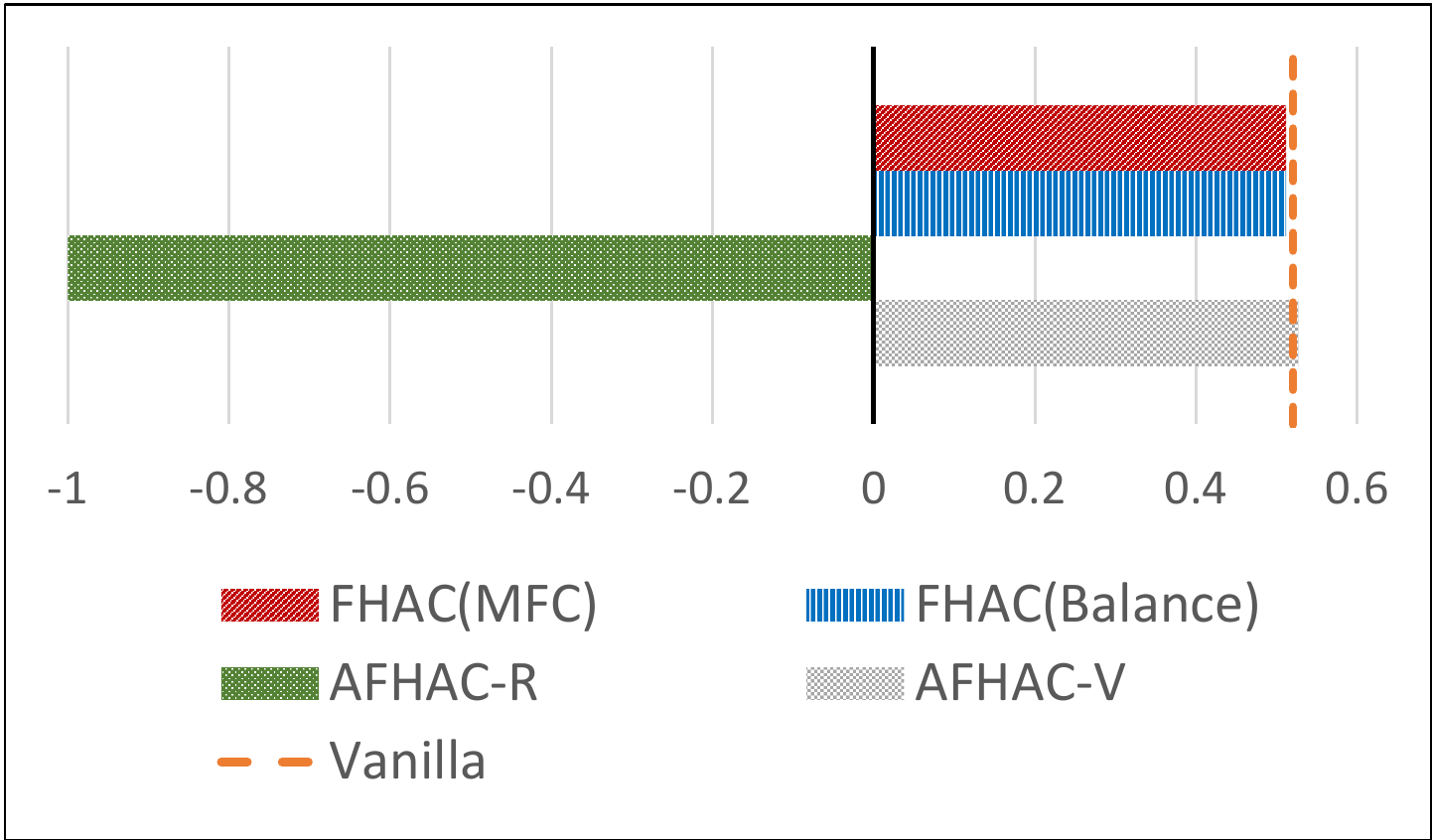}}
      \caption{Silhouette Scores for \texttt{census}}
\end{subfigure}
\caption{Clustering Performance Analysis}
\end{figure*}

In this subsection, we compare the performance of our proposed algorithm FHAC with AFHAC using the fairness metrics MFC and balance defined in Section \ref{section3}. AFHAC works with the following hierarchical clustering objectives: Cohen-Addad et al's \textit{value} objective \cite{cohen2019hierarchical} and Moseley et al's \textit{revenue} objective \cite{moseley2017approximation}. Experiments were run to calculate MFC and balance of the clusters formed by AFHAC optimizing revenue (AFHAC-R) and value (AFHAC-V). 

The AFHAC algorithm proposed by Ahmadian et al utilizes average-linkage HAC as part of their clustering process \cite{ahmadian2020fair}. Thus, for comparisons to be justifiable, we compare their algorithm with our proposed FHAC algorithm with average-linkage as the linkage criterion. Another important consideration is the difference in fairness enforcement. The AFHAC algorithms aim to ensure that all clusters (and their merged sub-clusters) should have each protected group's proportion upper-bounded by some value $\alpha$ provided at runtime. Extrapolating for the level with $k$ clusters, it is then evident that their algorithms cannot accommodate different \textit{ideal proportions} for each group as they attempt to use only the same $\alpha$ for each group. Mathematically, they aim to ensure that $\delta_g^{C_i} \leq \alpha$, $\forall g \in F, i\in [k]$. Therefore, to make comparisons fair, we set $\alpha = \max_{g \in F}\{\phi_g\}$ for our experiments using their algorithms. 

\begin{table}[htbp]
\caption{(MFC Results) FHAC, AFHAC-R, and AFHAC-V}\label{table3}
\begin{center}
    \tabcolsep=0.10cm
    \resizebox{0.42\textwidth}{!}{
    \begin{tabular}{|c|c|c|c|}
        \hline
    \textbf{Dataset}  & \textbf{MFC (FHAC)} & \textbf{MFC (AFHAC-V)} & \textbf{MFC (AFHAC-R)}\\
    \hline
    \hline
    \texttt{creditcard} & \textbf{1.0} & 2.8505 & 1.5825 \\ 
    \hline
    \texttt{census} & \textbf{0.0853} & 4.0 & 4.0 \\
    \hline
    \texttt{bank} & 0.6679 & \textbf{0.03648} & 2.0027\\
    \hline
    \end{tabular}}
\end{center}
\end{table}
\vspace*{-3mm}

We calculated MFC values for AFHAC-R and AFHAC-V on all the datasets mentioned in Section \ref{section4_1} and compare with Algorithm 1. These results are shown in Table \ref{table3}. We do similar experiments for balance and show the results in Table \ref{table4}. Observing both Table \ref{table3} and Table \ref{table4}, we can see that we outperform both AFHAC-V and AFHAC-R for both the MFC and balance metrics for the \texttt{creditcard} and \texttt{census} datasets. For the \texttt{bank} dataset, we outperform AFHAC-R for both MFC and balance, but AFHAC-V obtains better fairness utility. Despite this, we believe in the merits of our approach due to the lack of generalization capability of the AFHAC algorithms to arbitrary fairness notions, and their overall inconsistent performance on the other datasets.

\begin{table}[htbp]
\caption{(Balance Results) FHAC, AFHAC-R, AFHAC-V}\label{table4}
\begin{center}
    \tabcolsep=0.09cm
    \resizebox{0.42\textwidth}{!}{
    \begin{tabular}{|c|c|c|c|}
        \hline
    \textbf{Dataset} & \textbf{Balance(FHAC)} & \textbf{Balance(AFHAC-V)} & \textbf{Balance(AFHAC-R)}\\
    \hline
    \hline
    \texttt{creditcard} & \textbf{0.427} & 0.0 & 0.416 \\ 
    \hline
    \texttt{census} & \textbf{0.8865} & 0.2 & 0.0 \\
    \hline
    \texttt{bank} & 0.5567 & \textbf{0.9474} & 0.0\\
    \hline
    \end{tabular}}
\end{center}
\end{table}
\vspace*{-3mm}

\subsection{Comparing Clustering Performance of Fair Clusters}\label{section4_4}

\looseness-1 As a result of enforcing fairness constraints on a clustering, we are reducing the clustering quality compared to the original optimal clustering. This happens because we opt for more fair cluster merges over ones that minimize distance between clusters in Algorithm 1. That is, as we are improving fairness (either in terms of balance or MFC) for our proposed FHAC algorithm, we are reducing clustering performance compared to vanilla HAC. In our approach, this directly relates to increasing the $\beta$ relaxation of the distance criterion while finding pairs of clusters to merge together. Similarly, even for the AFHAC algorithms (and other fair algorithmic variants), clustering performance decreases at the cost of finding \textit{fairer} solutions.

Thus, we wish to analyze the clustering performance of our FHAC algorithm to the AFHAC algorithms. Considering the vanilla HAC clustering performance as the optimal, we obtain results to observe the extent to which clustering performance worsens and how it compares to the AFHAC algorithms. To measure clustering quality, we utilize the Silhouette Coefficient/Score \cite{ROUSSEEUW198753} which is a widely used clustering performance metric. It aims to capture intra-cluster similarity and inter-cluster dissimilarity and gives an output between $[-1,1]$ for a given clustering as input. A score of: $-1$ indicates an \textit{incorrectly assigned} clustering, $1$ indicates a \textit{dense and well-separated} clustering, and $0$ indicates \textit{overlapping clusters}. We use the Silhouette Score as it is easier to interpret compared to other \textit{unbounded} cluster performance metrics.


Since we are comparing Silhouette Coefficients for fair clusters obtained via Algorithm 1 and the AFHAC algorithms, it does not make sense to compare clustering solutions that are explicitly \textit{unfair}, i.e., solutions that have balance equal to $0$. Therefore, for any solutions such as these, we denote their Silhouette Score as $-1$ (incorrect clustering). The results comparing the Silhouette Scores for Algorithm 1 with average-linkage for MFC (denoted as FHAC(MFC)), Algorithm 1 with average-linkage for balance (denoted as FHAC(Balance)), AFHAC-V, and AFHAC-R, on the \texttt{creditcard}, \texttt{bank}, and \texttt{census} datasets are shown in Figure 2(a), Figure 2(b), and Figure 2(c), respectively. Quite evidently, FHAC achieves competitive clustering performance to the other algorithms.

\section{Conclusion}\label{section5}
\looseness-1 In this paper, we have proposed the FHAC algorithm (Algorithm 1) for performing HAC (Section \ref{section3}). Our algorithm works for multiple protected groups, generalizes to natural fairness notions for HAC, and for any linkage criterion used. We provide results on UCI datasets such as \texttt{bank}, \texttt{creditcard}, and \texttt{census} comparing our approach to vanilla HAC as well as the only other fair hierarchical clustering approach of \cite{ahmadian2020fair}, for both the MFC fairness cost and the balance fairness metric of \cite{chierichetti2017fair} (Section \ref{section4}). The results demonstrate that our proposed algorithm achieves better fairness and is more robust than the other clustering approaches. We further compare the clustering performance of our approach and show that it outputs clusters of quality comparable to the optimal as well as other related approaches. For future work, alternate approaches to making HAC fair (for e.g. post-processing approaches) can be explored.




 \bibliographystyle{IEEEtran}



\end{document}